\documentclass[runningheads]{llncs}
\usepackage[T1]{fontenc}

\usepackage{graphicx}
\usepackage{amsmath}
\usepackage{amssymb}
\begin{document}
\title{Less Traffic, Better Outcomes: Competition-Aware Request Dispatch in Real-Time Ad Exchanges}
\titlerunning{Less Traffic, Better Outcomes}
%
\author{
Jonaid Shianifar\inst{1}\orcidID{0000-0003-0477-0056} \and
Blaz Mramor\inst{1}\orcidID{0009-0001-9669-5374} \and
Fangda Zou\inst{1}\orcidID{0009-0007-4556-5768} \and
Matthieu C. Martin\inst{1}\orcidID{0009-0000-4751-1666} \and
Xingsheng Guo\inst{1}\orcidID{0009-0001-9198-5540} \and
Zhihua Zhu\inst{2}\orcidID{0000-0001-5702-4626} \and
Rong Zhou\inst{2}\orcidID{0000-0002-8764-5797} \and
Bichen Shi\inst{1}\orcidID{0000-0002-2965-3449}%
\thanks{Corresponding author.}
}
\authorrunning{J. Shianifar et al.}
%
\institute{
Huawei Ireland Research Center, Dublin, Ireland \\
\email{
jonaid.shianifar@huawei.com,
blaz.mramor@h-partners.com,
fangdazou@h-partners.com,
matthieu.c.martin@h-partners.com,
xingsheng.guo2@huawei.com,
bichen.shi@huawei-partners.com
}
\and
Huawei, Nanjing, China \\
\email{zhuzhihua8@huawei.com,
joe.zhourong@huawei.com}
}
\maketitle              
\begin{abstract}
Real-time bidding (RTB) ad exchanges typically forward nearly all incoming requests to demand-side platforms (DSPs), even though only a small fraction receive bids. This over-distribution weakens auction outcomes: DSPs throttle participation under compute and budget constraints, reducing the effective use of limited bidding capacity.
We present a competition-aware request dispatch framework that uses distributional bid prediction and probabilistic forwarding to decide whether each request should be sent to each DSP. The system adapts per-DSP thresholds over time through lightweight policy optimization to track non-stationary market conditions.
We evaluate the framework through four sequential online experiments on a production platform serving over 20 billion daily requests. A full multi-DSP deployment reduces DSP request volume under the policy by 34.2\% while increasing net revenue by 4.6\% ($p<0.001$) in a recent 14-day window after an initial DSP adaptation period.
Further analysis highlights strong heterogeneity across traffic segments and reveals that aggregate metrics can be misleading. Segment-level and per-DSP analyses suggest that the policy surfaces comparative advantages among DSPs, improving monetized outcomes without increasing overall request volume.

\end{abstract}

\section{Introduction}

Programmatic advertising relies on real-time bidding (RTB) auctions to allocate impressions at millisecond timescales. Ad exchanges receive impression opportunities from publishers and forward bid requests to demand-side platforms (DSPs), which decide whether and how much to bid. The dominant industry practice is to forward nearly all eligible requests under the assumption that more traffic increases auction revenue. In our production environment, however, fewer than 40\% of forwarded requests receive a bid.

This over-distribution creates inefficiencies beyond wasted infrastructure. DSPs operate under compute, latency, and budget constraints, and excessive low-value traffic can trigger throttling or selective participation. Reduced participation can degrade the allocation of limited DSP bidding capacity, lowering monetized outcomes. Prior work has shown that bidder participation is a key determinant of auction efficiency and pricing dynamics \cite{bulow1996auctions}.

The problem has become increasingly important due to the growth of header bidding, parallel auction architectures, and supply-path optimization (SPO). As request volumes rise and DSP inference pipelines become more expensive, exchanges that maximize raw traffic volume risk degrading marketplace efficiency and long-term demand quality.

Existing exchange-side approaches such as response-rate filtering and predictive throttling primarily optimize whether a DSP is likely to respond, rather than whether forwarding a request is likely to improve the auction itself. A DSP may respond while contributing little competitive value to the final outcome.

In this work, we propose a competition-aware request dispatch framework that explicitly models the value of bidder participation. For each request--DSP pair, the system estimates both the probability of receiving a bid and the conditional distribution of bid values. These signals are combined into a probabilistic forwarding policy that selectively dispatches requests to DSPs with stronger expected marginal contribution to auction outcomes. To adapt to non-stationary marketplace dynamics, per-DSP dispatch thresholds are periodically updated using lightweight reinforcement learning optimization on recent auction logs.

The framework is deployed on a production RTB platform serving more than 20 billion requests per day with end-to-end dispatch latency below 7ms on CPU infrastructure. We evaluate it through four sequential online experiments and provide stratified analyses revealing that aggregate auction metrics can mask important marketplace dynamics.

The main contributions of this paper are:

\begin{itemize}
    \item We formulate exchange-side request dispatch as a competition-aware optimization problem that estimates whether forwarding a request is likely to improve auction participation quality rather than simply maximize response rate.
    
    \item We develop a practical production framework combining distributional bid modeling, probabilistic request dispatch, and adaptive per-DSP threshold optimization under strict real-time serving constraints.
    
    \item We present large-scale online results from a production deployment serving over 20 billion daily requests, demonstrating that selective dispatch can simultaneously reduce DSP traffic volume and improve monetized outcomes.
    
    \item We provide stratified and per-DSP analyses showing that dispatch surfaces comparative advantages among DSPs, reshaping the competitive landscape in ways that aggregate metrics alone cannot capture.
\end{itemize}

\section{Related Work}

Research on real-time bidding (RTB) systems has primarily focused on advertiser-side bidding, budget pacing, bid landscape modeling, and auction optimization. Reinforcement learning and optimization methods have been widely applied to bidding and pacing under budget and market constraints \cite{zhang2014optimal,cai2017real,wu2018budget,agarwal2014budget,xu2015smart,wei2023rltp}. These approaches generally treat available auction opportunities as fixed, whereas our work studies how exchanges should control which opportunities are exposed to bidders.

Another line of work studies bid landscape and market-price forecasting for advertiser decision making. Distributional models have shown strong performance in estimating auction outcomes under uncertainty \cite{ren2019deep}. We use similar modeling ideas for a different objective: estimating whether forwarding a request is likely to contribute value to the auction outcome rather than predicting how much a bidder should bid.

On the supply side, prior work has explored impression allocation, multi-source delivery systems, and auction mechanism optimization \cite{wu2018multi,liao2022cross,pachilakis2019no,hu2025learning,fan2025two}. These methods optimize allocation, reserve pricing, or auction sequencing after bidder participation has already been determined. Our setting is complementary: we optimize an earlier stage by deciding which DSPs should receive each request.

Compared with prior work on bidding, pacing, and auction design, relatively less attention has been given to request forwarding itself as an exchange-side marketplace optimization problem. We position request dispatch as a competition-aware traffic curation mechanism that explicitly manages the quality--quantity trade-off in RTB systems.

\section{Methodology}

\begin{figure}[t]
    \centering
    \includegraphics[width=\linewidth]{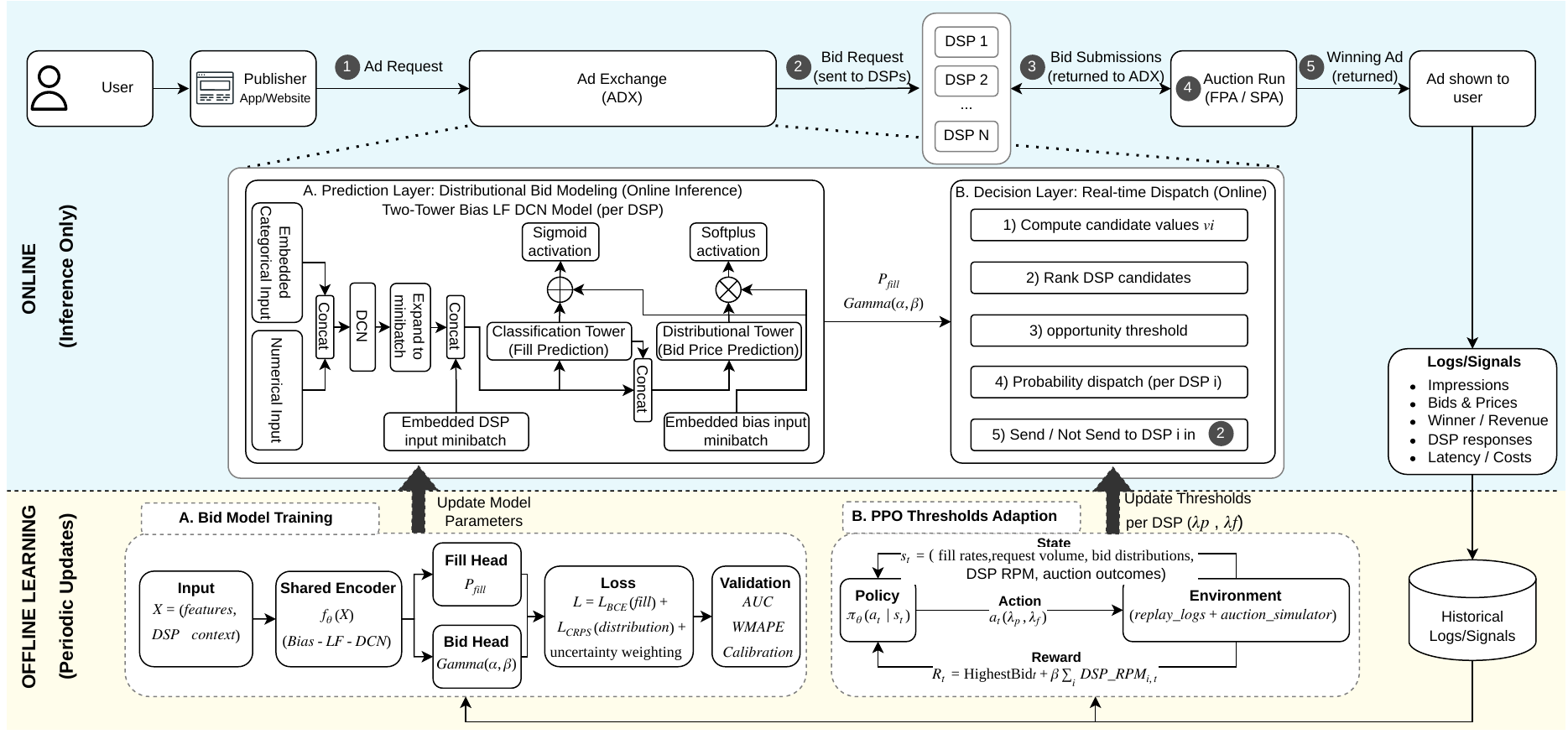}
    \caption{Competition-aware dispatch framework: prediction models estimate fill probability and bid value, the online layer sets forwarding probabilities, and an offline PPO loop updates per-DSP thresholds using auction and DSP response logs.}
    \label{fig:framework}
\end{figure}

Figure~\ref{fig:framework} illustrates the proposed competition-aware request dispatch framework. For each impression opportunity, the exchange estimates bid-response signals for eligible DSPs and selectively forwards requests using (i) distributional bid modeling, (ii) probabilistic competition-aware dispatch, and (iii) adaptive threshold optimization.

The online serving path consists only of lightweight inference and dispatch sampling, while model retraining and threshold optimization are executed asynchronously offline using recent auction logs. In production, the system serves more than 20 billion daily requests with end-to-end dispatch latency below 7\,ms on CPU infrastructure.

\subsection{Distributional Bid Modeling}

The routing controller relies on a production bid-response model to estimate both the probability that a DSP returns a bid and the conditional distribution of bid values on filled requests. For a candidate DSP $i$, we model:

\begin{equation}
p_i^{\mathrm{fill}}(X) = P(Y_{\mathrm{fill},i}=1 \mid X)
\end{equation}
\begin{equation}
Y_{\mathrm{bid},i} \mid Y_{\mathrm{fill},i}=1, X \sim \mathrm{Gamma}(\alpha_i,\beta_i)
\end{equation}
where $Y_{\mathrm{fill},i}$ indicates whether DSP $i$ returns a bid and $Y_{\mathrm{bid},i}$ denotes the bid value conditional on a fill event.

Our production model, denoted Bias-LF-DCN, is implemented in TensorFlow using a Deep \& Cross Network (DCN) backbone \cite{wang2017deep,wang2021dcn}. High-cardinality categorical features are embedded into dense vectors and concatenated with numerical context signals. The shared representation feeds two task-specific towers: a fill prediction tower and a bid distribution tower that outputs Gamma distribution parameters $(\alpha,\beta)$ through softplus activations.

To reduce production serving cost, the model uses late DSP fusion: shared request features are encoded once and combined with a minibatch of DSP-specific features only before the task towers. Compared with running a full model independently per DSP, this architecture substantially reduces CPU usage while preserving prediction quality.

\begin{table}[t]
\centering
\caption{Fusion strategies for multi-DSP serving.}
\label{tab:multi_dsp_fusion2}
\footnotesize
\begin{tabular}{p{0.35\linewidth}cc}
\hline
\textbf{Strategy} & \textbf{Compute Resource} & \textbf{Prediction Quality} \\
\hline
Per-DSP Bias-DCN (baseline) & 33\% & Reference \\
Fuse after embeddings & 32\% & Same \\
Fuse after DCN, before towers & 28\% & Comparable \\
Fuse before output heads & 22\% & Degraded \\
\hline
\end{tabular}
\end{table}

Table~\ref{tab:multi_dsp_fusion2} shows that fusion before the task towers substantially reduces CPU usage while preserving prediction quality. We therefore deploy this architecture in production.

The model is trained end-to-end with a multi-task objective combining binary cross-entropy for fill prediction and Continuous Ranked Probability Score (CRPS) \cite{GneitingRaftery2007} for bid distribution forecasting using homoscedastic uncertainty weighting\cite{kendall2018multi}. To adapt to market non-stationarity, the model is retrained daily on a rolling seven-day window.

\begin{table}[t]
\centering
\caption{Offline bid-model comparison.}
\label{tab:offline_model}
\small
\begin{tabular}{lcc}
\hline
Model & AUC $\uparrow$ & WMAPE $\downarrow$ \\
\hline
Simple baseline & $0.9600 \pm 0.0003$ & $0.4256 \pm 0.0115$ \\
Bias-LF-DCN & $0.9599 \pm 0.0001$ & $0.4064 \pm 0.0062$ \\
Bias-DCN & $\mathbf{0.9613 \pm 0.0002}$ & $\mathbf{0.4005 \pm 0.0044}$ \\
\hline
\end{tabular}
\end{table}

Bias-DCN achieves the best offline prediction accuracy, while Bias-LF-DCN provides a substantially better compute--performance trade-off for production serving. We therefore deploy Bias-LF-DCN in the online system.

\subsection{Competition-Aware Dispatch}

The objective of dispatch is not merely to maximize response rate, but to forward requests where a DSP has high expected marginal contribution to auction outcomes. For each RTB DSP $i$, we compute an opportunity value:

\begin{equation}
v_i^{(\mathrm{rtb})}
=
p_i^{\mathrm{fill}}
\cdot
\mathbb{E}[Y_{\mathrm{bid},i}\mid Y_{\mathrm{fill},i}=1,X]
=
p_i^{\mathrm{fill}}
\cdot
\frac{\alpha_i}{\beta_i}
\end{equation}

The exchange also considers programmatic guaranteed demand. Let $\mathcal{V}$ denote the set of all RTB and guaranteed-demand opportunity values, sorted in descending order. The competition threshold is defined as:

\begin{equation}
\tau = v_{(K)}
\end{equation}
where $K$ is the maximum number of ads that can be displayed. 
Because $\tau$ depends on the opportunity values of all candidate DSPs and guaranteed-demand sources for the current request, the forwarding decision for each DSP implicitly conditions on the competitive context it faces in this auction.
For each DSP, the system estimates the probability that its bid exceeds the adjusted competition threshold:

\begin{equation}
p_i^{\mathrm{comp}}
=
P(Y_{\mathrm{bid},i}>\lambda_p^{(i)}\tau
\mid
Y_{\mathrm{fill},i}=1,X)
\end{equation}

where $\lambda_p^{(i)}$ is a per-DSP conservativeness parameter.

A smooth fill gate additionally suppresses low-response traffic:

\begin{equation}
G_i^{\mathrm{fill}}
=
\sigma\!\left(
\operatorname{logit}(p_i^{\mathrm{fill}})
-
\operatorname{logit}(\lambda_f^{(i)})
\right)
\end{equation}

The final forwarding probability is:

\begin{equation}
p_i^{\mathrm{fwd}}
=
p_{\min}^{(i)}
+
(1-p_{\min}^{(i)})
\,
p_i^{\mathrm{comp}}
\,
G_i^{\mathrm{fill}}
\end{equation}
where $p_{\min}^{(i)}$ is a small exploration floor that preserves counterfactual coverage and reduces selection bias. The exchange then samples the forwarding action:

\begin{equation}
a_i \sim \mathrm{Bernoulli}(p_i^{\mathrm{fwd}})
\end{equation}

This probabilistic dispatch mechanism avoids sharp threshold boundaries while maintaining exploration for future retraining.

\subsection{PPO-based Adaptive Threshold Optimization}

Fixed dispatch thresholds are insufficient because DSP behavior changes over time and thresholds are coupled through marketplace competition. Changing one DSP's forwarding threshold can alter its request mix, response rate, bidding behavior, pacing, and participation, which in turn affects the competitive context faced by other DSPs. These effects may also be delayed due to DSP budget, throttling, and capacity constraints. To adapt to these non-stationary dynamics, the system periodically updates per-DSP dispatch thresholds offline.

At update step $t$, the Proximal Policy Optimization (PPO) state summarizes recent marketplace statistics aggregated over the latest serving windows:
\begin{equation}
\begin{split}
s_t =
\{
\text{fill rates},
\text{request volume},
\text{bid distributions},\\
\text{DSP RPM},
\text{auction outcomes}
\}
\end{split}
\end{equation}

The action corresponds to updating the per-DSP dispatch parameters:
\begin{equation}
a_t =
\{
\lambda_p^{(i)},
\lambda_f^{(i)}
\}_{i=1}^{N}
\end{equation}
where $\lambda_p^{(i)}$ controls competition conservativeness and $\lambda_f^{(i)}$ controls low-fill filtering aggressiveness for DSP $i$, and $N$ is the total number of DSPs. The reward balances auction value and request efficiency:
\begin{equation}
R_t
=
\mathrm{HighestBid}_t
+
\beta
\sum_i
\mathrm{DSP\_RPM}_{i,t}
\end{equation}
where $\mathrm{DSP\_RPM}$ denotes revenue per thousand DSP requests and $\beta$ controls the trade-off between auction-level value and request efficiency. $\beta$ was selected empirically in the experiments.

We therefore formulate threshold adaptation as a sequential decision-making problem and implement it using PPO \cite{schulman2017ppo}. PPO is used to update thresholds jointly using aggregated marketplace statistics, while the online serving system only reads periodically updated threshold tables, keeping dispatch deterministic and low-latency.

\section{Experimental Setup}

We evaluate the proposed framework through both offline simulation and online production experiments on a large-scale RTB platform serving more than 20 billion requests per day. The experiments are designed to measure whether selective dispatch can improve marketplace efficiency and monetized outcomes while reducing DSP request volume.

\subsection{Offline Validation and Ablation}

We first compare dispatch strategies in an offline auction simulator constructed from replayed production logs. The simulator evaluates relative strategy ordering rather than exact marketplace reproduction, since downstream billing and long-term DSP behavior are only partially observable offline.

We compare four strategies:
\begin{itemize}
    \item \textbf{Full forwarding:} production baseline that forwards all eligible requests;
    \item \textbf{Random filtering:} random request suppression with matched traffic reduction;
    \item \textbf{Best static threshold:} fixed thresholds optimized offline;
    \item \textbf{PPO adaptation (ours):} adaptive threshold optimization using PPO.
\end{itemize}

Table~\ref{tab:offline_compare} summarizes the offline comparison. 
Random filtering substantially degrades auction quality, while static thresholds achieve only limited request reduction before harming outcomes. PPO-based adaptation preserves or slightly improves highest bid while reducing request volume by approximately 35\%. 
The best static strategy represents exhaustive offline search yet achieves only half the request reduction, confirming that adaptive optimization provides value beyond fixed tuning.

\begin{table}[t]
\centering
\caption{Offline strategy comparison.}
\label{tab:offline_compare}
\small
\begin{tabular}{lccc}
\hline
Method & Highest Bid & DSP Requests & DSP RPM \\
\hline
Full forwarding & baseline & baseline & baseline \\
Random ($p{=}0.65$) & $-$35\% & $-$35\% & $0\%$ \\
Best static & $-$1.8\% & $-$17.5\% & +19.0\% \\
PPO (ours) & +1.5\% & $-$35\% & +56.1\% \\
\hline
\end{tabular}
\end{table}

\subsection{Production Deployment and Evaluation}

The framework was evaluated through four sequential online experiments. E1--E3 deploy the policy on individual DSPs within the Mid-RPM traffic stratum, while E4 applies the policy simultaneously across the top-$N$ DSPs (covering more than 80\% of total DSP traffic) on full inventory. E4 uses a 7-day pre-period and 20-day post-period.

Standard A/B testing is challenging in RTB systems due to DSP budget interference, temporal demand drift, and delayed behavioral adaptation. We therefore use a Ratio Difference-in-Differences (Ratio-DID) estimator. Let

\begin{equation}
R_{\mathrm{pre}}
=
\frac{\bar{Y}_{\mathrm{pre}}^D}
{\bar{Y}_{\mathrm{pre}}^C},
\qquad
R_{\mathrm{post}}
=
\frac{\bar{Y}_{\mathrm{post}}^D}
{\bar{Y}_{\mathrm{post}}^C}
\end{equation}
denote treatment-to-control ratios in the pre- and post-periods. The Ratio-DID estimator is:

\begin{equation}
\mathrm{Ratio\text{-}DID}
=
\frac{R_{\mathrm{post}}}
{R_{\mathrm{pre}}}
-1
\end{equation}

This formulation normalizes persistent level differences between groups and removes common multiplicative time trends under a parallel-trend assumption. Statistical significance is computed using permutation tests with 10{,}000 iterations.

For the multi-DSP deployment (E4), traffic was partitioned into four persistent hash buckets at the slot level. Bucket D received the dispatch policy, while buckets A+B+C remained under the production baseline and were pooled as controls.

\begin{table}[t]
\centering
\caption{Pre-period stability diagnostics for E4 (D vs.\ A+B+C).}
\label{tab:pre_stability}
\small
\begin{tabular}{lrrr}
\hline
Metric & Pre CV & Pre drift \\
\hline
Net revenue & 2.13\% & $-$1.1\%  \\
DSP requests (sent) & 6.01\% & $-$9.1\%  \\
DSP responses & 3.80\% & $-$3.1\%  \\
Highest Bid & 3.38\% & $-$1.6\%  \\
\hline
\end{tabular}
\end{table}

Pre-period diagnostics are summarized in Table~\ref{tab:pre_stability}. Net revenue shows relatively stable treatment-to-control ratios. Request-volume metrics exhibit larger pre-period drift but remain small relative to expected treatment effects, so they do not compromise causal interpretation.

We report three classes of metrics:
\begin{itemize}
    \item \textbf{Revenue metrics:} net revenue and highest bid;
    \item \textbf{Efficiency metrics:} fill rate, impression rate, and DSP RPM;
    \item \textbf{Volume metrics:} DSP requests, responses, impressions, and clicks.
\end{itemize}

The production deployment additionally includes several operational guardrails. High-value requests retain a minimum forwarding probability, threshold updates are bounded between update windows, and automated rollback is triggered if revenue, response rate, or latency metrics cross predefined safety thresholds.

\section{Results and Discussion}

\subsection{Single-DSP Validation}

Table~\ref{tab:e1e3} summarizes three sequential experiments on a single DSP within the Mid-RPM stratum, each with a 7-day pre-period and 14-day post-period applied to different slot partitions. Across all rounds, selective dispatch reduces request volume by 34--71\% while consistently improving fill rate and DSP RPM.

\begin{table}[t]
\centering
\caption{Single-DSP Ratio-DID results (Mid-RPM stratum).}
\label{tab:e1e3}
\small
\begin{tabular}{lccccccc}
\hline
Exp & Net Rev. & DSP Req. & DSP Resp. & Highest Bid & Fill Rate & RPM \\
\hline
E1 & +24.3\%\textsuperscript{*} & $-$34.1\%\textsuperscript{*} & +77.4\%\textsuperscript{***} & +53.0\%\textsuperscript{***} & +169.4\% & +88.6\%\textsuperscript{*} \\
E2 & +15.1\%\textsuperscript{*} & $-$70.8\%\textsuperscript{*} & +15.0\% & +15.6\% & +293.7\% & +294.2\%\textsuperscript{*} \\
E3 & $-$1.2\% & $-$40.8\%\textsuperscript{***} & +13.6\% & +5.1\% & +91.7\% & +66.7\% \\
\hline
\multicolumn{7}{l}{\footnotesize \textsuperscript{***}\,$p<0.001$,\; \textsuperscript{*}\,$p<0.05$}
\end{tabular}
\end{table}

Highest bid increases in all three experiments, while raw bid eCPM remains approximately stable, suggesting that request reduction does not systematically suppress bidding. E1 and E2 show strong revenue gains, while E3 is approximately revenue-neutral despite large efficiency improvements. This suggests that monetization gains depend on DSP- and inventory-specific demand, not request reduction alone.

\subsection{Multi-DSP Deployment}

E4 extends the policy from single-DSP Mid-RPM traffic to full-inventory multi-DSP dispatch. This setting adds two sources of heterogeneity: Low-RPM traffic dominates volume but has low impression realization, and simultaneous DSP updates create interaction effects. We therefore report aggregate results together with stratified diagnostics.

\begin{figure}[t]
    \centering
    \includegraphics[width=\linewidth]{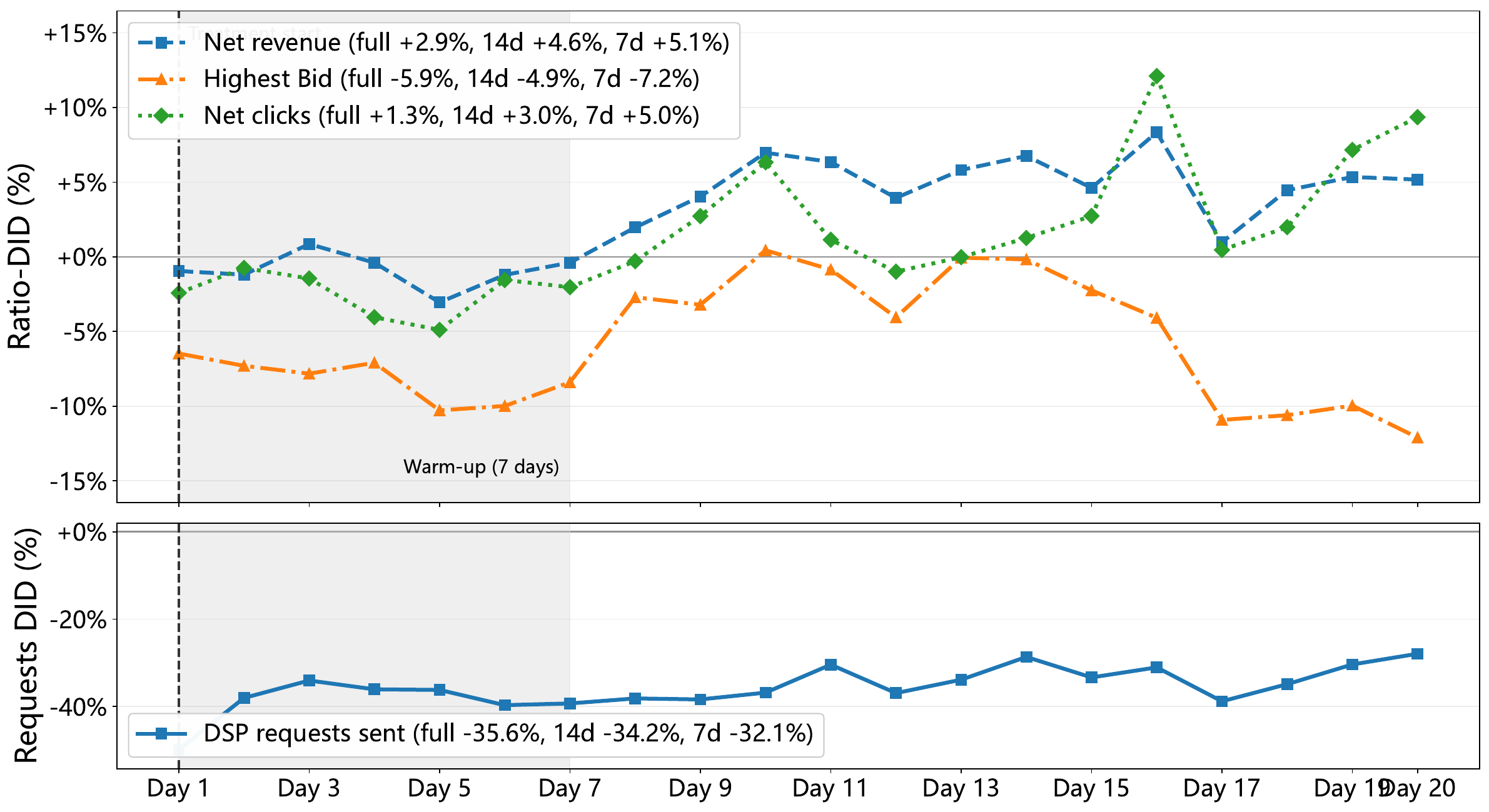}
    \caption{Daily Ratio-DID trajectories (D vs A+B+C) for E4. Dashed line: policy activation. Shaded region: initial adaptation window.}
    \label{fig:e4daily}
\end{figure}

Table~\ref{tab:e4agg} reports aggregate Ratio-DID results using pooled controls (A+B+C). We separately report recent post-adaptation windows because the first week exhibits observable threshold and DSP-response adaptation (Figure~\ref{fig:e4daily}).

\begin{table}[t]
\centering
\caption{E4 aggregate Ratio-DID (D vs.\ A+B+C).}
\label{tab:e4agg}
\small
\begin{tabular}{lccc}
\hline
Metric & Full 20d & Recent 14d & Recent 7d \\
\hline
Net revenue & +2.9\%\textsuperscript{**} & +4.6\%\textsuperscript{***} & +5.1\%\textsuperscript{*} \\
DSP requests & $-$35.6\%\textsuperscript{***} & $-$34.2\%\textsuperscript{***} & $-$32.1\%\textsuperscript{*} \\
DSP responses & $-$7.3\%\textsuperscript{***} & $-$6.7\%\textsuperscript{***} & $-$8.2\%\textsuperscript{*} \\
Highest Bid & $-$5.9\%\textsuperscript{***} & $-$4.9\%\textsuperscript{***} & $-$7.2\%\textsuperscript{*} \\
Fill rate & +43.9\%\textsuperscript{***} & +41.8\%\textsuperscript{***} & +35.2\%\textsuperscript{***} \\
RPM & +59.8\%\textsuperscript{***} & +59.0\%\textsuperscript{***} & +54.8\%\textsuperscript{***} \\
Net clicks & +1.3\% & +3.0\%\textsuperscript{*} & +5.0\%\textsuperscript{*} \\
Impressions & $-$1.8\%\textsuperscript{***} & $-$1.5\%\textsuperscript{*} & $-$1.5\% \\
\hline
\multicolumn{4}{l}{\footnotesize \textsuperscript{***}\,$p<0.001$,\; \textsuperscript{**}\,$p<0.01$,\; \textsuperscript{*}\,$p<0.05$}
\end{tabular}
\end{table}

Table~\ref{tab:e4agg} shows that E4 reduces DSP requests by 34.2\% and increases net revenue by 4.6\% ($p<0.001$) in the recent 14-day window. DSP responses decline only 6.7\%, so fill rate rises by more than 40\%. 
Net clicks increase by 3.0\% despite a slight decline in impressions ($-$1.5\%), indicating that the policy retains higher-quality impressions that are more likely to convert.
The negative aggregate highest-bid effect is expected: the policy primarily suppresses Low-RPM requests whose impression realization rate is only 4.9\% (Table~\ref{tab:baseline}), so internal bid changes on this traffic have minimal revenue transmission. The stratified analysis below confirms that revenue-producing strata show positive highest-bid effects.

\subsection{Stratified and Per-DSP Analysis}

\begin{table}[t]
\centering
\caption{Baseline properties by stratum.}
\label{tab:baseline}
\small
\begin{tabular}{lccc}
\hline
Property & High-RPM & Mid-RPM & Low-RPM \\
\hline
Net revenue share & 13.5\% & 23.0\% & 63.5\% \\
Request share & 0.7\% & 2.8\% & 96.4\% \\
RPM & 1.062 & 0.482 & 0.038 \\
Impression rate & 34.0\% & 36.0\% & 4.9\% \\
\hline
\end{tabular}
\end{table}

\begin{table}[t]
\centering
\caption{Stratified Ratio-DID (recent 7-day window).}
\label{tab:strat}
\small
\begin{tabular}{lccc}
\hline
Metric & High-RPM & Mid-RPM & Low-RPM \\
\hline
Net revenue & +1.3\% & \textbf{+10.5\%}\textsuperscript{***} & $-$0.9\% \\
DSP requests & $-$18.8\%\textsuperscript{***} & $-$11.3\%\textsuperscript{***} & $-$28.3\%\textsuperscript{***} \\
Highest bid & $-$8.2\%\textsuperscript{**} & \textbf{+4.6\%} & $-$4.4\%\textsuperscript{***} \\
Fill rate & +7.8\%\textsuperscript{**} & +9.9\%\textsuperscript{***} & +32.2\%\textsuperscript{***} \\
Impr.\ rate & +15.3\%\textsuperscript{***} & +3.5\%\textsuperscript{***} & +6.7\%\textsuperscript{***} \\
RPM & +25.0\%\textsuperscript{***} & +23.7\%\textsuperscript{***} & +37.9\%\textsuperscript{***} \\
\hline
\multicolumn{4}{l}{\footnotesize \textsuperscript{***}\,$p<0.001$,\; \textsuperscript{**}\,$p<0.01$,\; \textsuperscript{*}\,$p<0.05$}
\end{tabular}
\end{table}

Tables~\ref{tab:baseline} and~\ref{tab:strat} explain the aggregate pattern. Low-RPM traffic accounts for 96.4\% of requests and drives the negative aggregate highest-bid effect, whereas Mid-RPM traffic produces the largest revenue lift (+10.5\%) and a positive highest-bid effect. High-RPM traffic changes little in revenue, consistent with limited headroom where DSP participation is already effective. 
Impression rate, which reflects how often an internal winning bid clears the external auction and renders, improves across all strata (Table~\ref{tab:strat}), indicating stronger external competitiveness of the bids retained under the policy.
Thus, aggregate internal bid price is not the right optimization objective: revenue depends on the full chain from internal bid to realized impression and external billing, and Mid-RPM traffic is where dispatch creates the most value.

\subsubsection{Comparative Advantage Across DSPs}

\begin{table}[t]
\centering
\caption{Per-DSP Mid-RPM auction funnel (recent 7-day window).}
\label{tab:funnel}
\small
\begin{tabular}{lcc}
\hline
Metric & DSP-A & DSP-B \\
\hline
Requests & $-$13.3\% & $-$4.2\% \\
Responses & +4.4\% & $-$2.2\% \\
Highest bid & +6.9\% & +6.7\% \\
Bid price & $-$0.8\% & \textbf{+7.6\%} \\
eCPC (rev/click) & $-$5.0\% & \textbf{+15.6\%} \\
Billing ratio & +0.3\% & +1.8\% \\
Net revenue & +7.4\% & \textbf{+18.1\%} \\
\hline
\end{tabular}
\end{table}

Table~\ref{tab:funnel} shows heterogeneous DSP responses. DSP-A follows a request-refinement pattern: fewer requests but more responses and higher highest bids at stable bid prices. DSP-B instead shows higher bid price, eCPC, billing ratio, and revenue. These patterns support the comparative-advantage view of dispatch: filtering low-value request--DSP pairs concentrates each DSP on traffic where its demand, models, or budgets are better aligned with the available impressions.

We do not causally isolate DSP strategy adaptation from traffic composition effects, so cross-DSP interaction patterns should be interpreted as diagnostic rather than definitive evidence. The consistent system-level result is nevertheless clear: large request reductions can coexist with improved monetized outcomes when dispatch optimizes participation quality rather than raw traffic volume.

\section{Conclusion}

We presented a competition-aware request dispatch framework for large-scale RTB ad exchanges. The system combines distributional bid modeling, probabilistic forwarding, and adaptive threshold optimization to selectively dispatch requests toward DSPs with stronger comparative advantage in expected auction outcomes, while operating under strict production latency constraints.

Across four online experiments, the framework consistently improved request efficiency while substantially reducing DSP traffic volume. In the full multi-DSP deployment, the policy reduced DSP requests by 34.2\% while improving net revenue by 4.6\% after an initial adaptation period. Stratified analysis revealed that aggregate auction metrics can mask important heterogeneity: the policy's primary value lies not in raising internal auction prices uniformly, but in surfacing comparative advantages among DSPs and improving the exchange's external competitiveness across the full monetization chain.

The current evaluation has several limitations: PPO adaptation operates on aggregated statistics rather than individual auctions, offline simulation only partially reproduces DSP behavior, and results are derived from a single exchange environment.

Nevertheless, the results suggest that traffic curation and participation quality can be more important than maximizing request volume alone in RTB marketplaces. Future work includes stronger causal identification of cross-DSP competition effects and longer-term modeling of DSP adaptation dynamics.

\begin{credits}

\subsubsection{\discintname}
The authors are affiliated with Huawei or Huawei Ireland Research Center. The authors declare that they have no other competing interests relevant to the content of this article.

\end{credits}
%
%
%
\bibliographystyle{splncs04}
\bibliography{bibliography}

\end{document}